%% file: emnlp2021.tex
\pdfoutput=1
\documentclass[11pt]{article}

\usepackage[]{emnlp2021}
\usepackage{times}
\usepackage{latexsym}
\usepackage{amssymb}
\usepackage{amsmath}
\usepackage{pifont}
\usepackage{adjustbox}

\usepackage[T1]{fontenc}
\usepackage[utf8]{inputenc}
\usepackage{microtype}

\usepackage{nameref}

\newcommand{\xmark}{\ding{55}}%
\newcommand{\qm}{{\fontfamily{cyklop}\selectfont \textit{?}}}

\usepackage{tabularx}

\newcolumntype{C}{>{\centering\arraybackslash}X}

\title{Sequence-to-Sequence Lexical Normalization with Multilingual Transformers}

\author{Ana-Maria Bucur\thanks{$^*$ Equal contribution} \\ University of Bucharest \\ ana-maria.bucur@drd.unibuc.ro
        \And
        Adrian Cosma\footnotemark[1] \\ University Politehnica of Bucharest \\ cosma.i.adrian@gmail.com
        \And
        Liviu P. Dinu \\ University of Bucharest \\ ldinu@fmi.unibuc.ro
}

\begin{document}
\maketitle
\begin{abstract}

Current benchmark tasks for natural language processing contain text that is qualitatively different from the text used in informal day to day digital communication. This discrepancy has led to severe performance degradation of state-of-the-art NLP models when fine-tuned on real-world data. One way to resolve this issue is through lexical normalization, which is the process of transforming non-standard text, usually from social media, into a more standardized form. In this work, we propose a sentence-level sequence-to-sequence model based on mBART, which frames the problem as a machine translation problem. As the noisy text is a pervasive problem across languages, not just English, we leverage the multi-lingual pre-training of mBART to fine-tune it to our data. While current approaches mainly operate at the word or subword level, we argue that this approach is straightforward from a technical standpoint and builds upon existing pre-trained transformer networks. Our results show that while word-level, intrinsic, performance evaluation is behind other methods, our model improves performance on extrinsic, downstream tasks through normalization compared to models operating on raw, unprocessed, social media text.

\end{abstract}

\section{Introduction}
\input{sections/1.Introduction}

\section{Related Work}
\input{sections/2.RelatedWork}

\section{Data and Evaluation}
\input{sections/3.DataandEvaluation}

\section{Method}
\label{method}
\input{sections/4.Method}

\section{Results}
\input{sections/5.Results}

\section{Conclusions}

In this work, we presented a method to perform lexical normalization by fine-tuning a multilingual machine translation model on pairs of noisy and normalized sentences from various languages. We employed mBART, as it is currently the state-of-the-art in transformer-based multilingual machine translation, allowing us to fine-tune on all available languages simultaneously. Furthermore, we used mBART as a denoising autoencoder and tuned it in a supervised fashion. 

As opposed to current two-stage methods for word candidate generation and ranking, our approach is more straightforward. Moreover, it scales to multiple languages without increasing computational demand (i.e. not increasing vocabulary size, increasing search space and others). Evaluation results show that our method, even though it lacks behind current methods on intrinsic, word-level evaluation, improves performance on downstream tasks. 

For future work, we aim to develop our method for better post-processing of the output and increasing augmentation levels - i.e. injecting more noise in the form of spelling mistakes, backwards translations etc. Moreover, since our method is supervised, the quality and quantity of training data play an essential role in the final performance. In this regard, we aim to explore ways to take into account inconsistent annotations.

\section*{Acknowledgments}

We would like to thank the reviewers for the insightful feedback provided. This research was partially supported by Blog Alchemy Limited.

\bibliography{custom}
\bibliographystyle{acl_natbib}

\end{document}

%% file: sections/1.Introduction.tex
Social media is a pervasive part of our modern lives and provides us with a rich source of information and insight into human behaviour. User-generated content has been a valuable resource for the research community, especially in the form of text, but it is notoriously noisy and non-standard. Models that operate on social media posts go beyond marketing and advertisement applications, and have the potential to impact real human lives through, for instance, detecting loneliness \cite{guntuku2019studying}, stress \cite{winata2018attention},  life satisfaction \cite{yang2016life}, suicidal ideation \cite{matero2019suicide, cao2019latent}, and mental health problems such as depression \cite{yates2017depression, bucur2021early, tadesse2019detection} and PTSD \cite{coppersmith2014measuring, amir2019mental}.

%%%%%%%%%%%%%%%%%%%%%%%%%%%%%%%%%%%%%%%%%%%%%%%%%%%%%%%%%%%%
Outside of a formal setting, users communicate freely in text form, resorting to abbreviations, slang or plain spelling mistakes or typos. \citet{eisenstein2013bad} further explored \textit{bad language} on social media, in the sense of language that defies our expectation of \textit{good} spelling, vocabulary and syntax. He identified several underlying factors for the cause of non-standard text: user illiteracy, length limits imposed by social media sites (i.e. Twitter), text input affordances (i.e. standard mobile keyboards or predictive entry), pragmatics (emoticons/emoji, abbreviations and expressive lengthening), and a social component. \citet{nguyen2021learning} further explored the latter, concluding that some types of non-standard text have strong social meaning, and normalization could induce a loss of meaning. 

However, it is well known that for most benchmark tasks, noisy/non-standard text has proven to be a real problem to NLP models, such as BERT \cite{kumar-etal-2020-noisy}, trained on clean or curated data, but fine-tuned on tasks with noisy and inconsistent format. 

%%%%%%%%%%%%%%%%%%%%%%%%%%%%%%%%%%
To overcome this predicament, \citet{eisenstein2013bad} proposes two possible approaches: either \textit{domain adaptation} or \textit{normalization}. While domain adaptation is not specific to natural language processing, text normalization and cleaning have always been a central part of any modern text processing pipeline. Text normalization is the process of adapting an input text to a more standard form. It has proven to be effective in increasing performance on tasks such as POS tagging  \cite{van-der-goot-cetinoglu-2021-lexical}, dependency parsing \cite{van2019depth} and sentiment analysis \cite{mandal2018normalization}. Naturally, most text normalization pipelines are based on supervised models, which require carefully annotated data. However, annotating a large corpus of text in multiple languages is often cumbersome and expensive, and some approaches rely on synthetically generating corrupted text \cite{dekker2020synthetic,ma2019nlpaug}.
%%%%%%%%%%%%%%%%%%%%%%%%%%%%%%%%%%

Commonly, approaches are based on word-level normalization. One of the most prominent methods is MoNoise \cite{van2017monoise}, in which the text correction pipeline is similar to a classic ranked retrieval. However, MoNoise operates at the individual word and uses a spelling correction module and a word embedding module. While word embeddings can be made to account for a specific sentence context, it is mostly discarded. 

Different from current methods, we aim to perform text normalization at a sentence level. This approach has several advantages, compared to word or subword methods: i) it can be naturally framed as a sequence-to-sequence type problem, ii) it is more straightforward, as it requires only one module, as opposed to a multi-stage pipeline (i.e. complex candidate generation and ranking), and iii) the same model can be trained on multiple languages at the same time, without increasing in size and computational processing.

In this edition of The  Workshop on  Noisy  User-generated  Text (W-NUT), organizers propose the shared task of multilingual lexical normalization\footnote{\url{http://noisy-text.github.io/2021/multi-lexnorm.html}}, in which participants are required to perform lexical normalization on 12 different languages \cite{multilexnorm}. 

As such, we use the state-of-the-art multilingual sequence-to-sequence transformer model mBART \cite{tang2020multilingual} and fine-tune it for our task. mBART is one of the first models that can be fine-tuned simultaneously on multiple languages without performance loss. We show that framing text normalization as a neural machine translation problem is a viable method for text normalization, improving performance on extrinsic, downstream tasks compared to models that operate on raw, unprocessed social media text. We made the code publicly available on github.\footnote{\url{https://github.com/bucuram/seq2seq-multilingual-normalization}}

%% file: sections/2.RelatedWork.tex
The W-NUT workshop hosted a shared task on lexical normalization of user-generated content from English tweets in its first edition \cite{baldwin2015shared}. The task received from the competing teams two categories of submissions, from constrained (using only the training data provided by the organizers) and unconstrained systems (using other publicly available data or tools). 

The best model, from \citet{jin2015ncsu}, generated candidates from the most similar canonical forms from the training data evaluated with the Jaccard Index. A random forest classifier was used to predict the suitable canonical form from all the candidates using features such as support and confidence, string similarity, and part of speech tags. The model was a constrained system, suggesting that the quality of the proposed model is more important than using additional data and tools. Other approaches were based on conditional random fields (CRF) \cite{akhtar2015iitp, supranovich2015ihs_rd, akhtar2015iitp} and recurrent neural networks (RNN) \cite{min2015ncsu_sas_wookhee, wagner2015dcu} among others.

Notably, MoNoise \cite{van2017monoise} has long been considered state-of-the-art in lexical normalization. MoNoise is a normalization model using spelling correction and word embeddings for candidate generation and a feature-based random forest classifier for candidate ranking. It is a modular normalization system easily reusable and adaptable \cite{van2017monoise}. The model was at the beginning developed only for English text. Still, then it was later expanded for multi-lingual lexical normalization covering languages such as Dutch, Spanish, Turkish, Slovenian, Croatian and Serbian \cite{van2019monoise}.

The lexical normalization task can also be formulated as a machine translation (MT) task. The noisy user-generated content is the source language, and the canonical form is the target language. \citet{veliz2019comparing} compare the MT approaches for lexical normalization, focusing on statistical neural translation (SMT) and neural machine translation (NMT) and obtaining better results using the SMT method. Furthermore, the authors show that the SMT approach works better in a low-resource setting than an NMT approach which requires a lot of data.

With the rise in popularity of pre-trained language models for natural language understanding and natural language generation, their ability to perform lexical normalization was also studied. By transforming the task into a token prediction one, \citet{muller2019enhancing} demonstrate that a BERT model can be used as a lexical normalization model in low resource settings.

Current methods for lexical normalization attempt to normalize at the character-level \cite{pennell2011character,ljubevsic2014standardizing}, syllable-level \cite{xu2015tweet}, word-level \cite{van2019monoise, jin2015ncsu} or sentence-level \cite{muller2019enhancing, lourentzou2019adapting}. \citet{lusetti2018encoder} propose an encoder-decoder approach for text normalization. 

We propose to make use of the latest transformer models that are capable of multilingual translation in a sequence to sequence manner, namely mBART \cite{tang2020multilingual}. However, we do not perform translation between languages, but instead, we use mBART as a denoising autoencoder, i.e. translating from \textit{bad English} to \textit{good English}. This way, we take the whole sentence into consideration when correcting the text. Moreover, this method is more straightforward and can scale to multiple languages without increasing computational demands.

%% file: sections/3.DataandEvaluation.tex
We further describe the dataset for this task and evaluation procedures.

\noindent \textbf{MultiLexNorm Dataset} The data provided by the organizers includes texts from 12 languages: Croatian, Danish, Dutch, English, German, Italian, Serbian, Slovenian, Spanish, Turkish and code-switched data for Indonesian-English and Turkish-German, as seen in Table \ref{tab:data}. Some examples from the training data are shown in Table \ref{tab:examples}. For some languages in the dataset, the capitalization (Caps column) is also corrected, and words are split or merged (1-N/N-1 column). The dataset comprises Twitter posts from all languages, but some languages also have texts from additional sources. For example, Danish also has texts from Arto, Denmark’s first large-scale social media \cite{plank-etal-2020-dan} and Dutch texts were also gathered from public Internet forums, and SMS messages \cite{dutchNorm}.

\begin{table}[hbt!]
    \centering
    \begin{adjustbox}{width=\linewidth}
    \begin{tabular}{l|rrrr} 
    \textbf{Language}  & \textbf{Words}  & \textbf{1-N/N-1} & \textbf{Caps} & \textbf{\%normed}\\
    \hline\hline
    Croatian &  75,276 & - & + & 8.98\\
    Danish & 11,816 & + & + & 8.66 \\
    Dutch & 23,053 & + & + & 26.49 \\
    English & 73,806 & + & - & 6.90 \\
    German &  25,157 & + & + & 8.90 \\
    Indonesian-English & 23,124 & + & - & 12.16 \\
    Italian & 14,641 & + & + & 7.36 \\
    Serbian & 91,738 & - & + & 7.73 \\
    Slovenian & 75,276 & - & + & 15.66 \\ 
    Spanish & 13,827 & - & - & 7.69 \\
    Turkish & 7,949 & - & + & 36.60 \\
    Turkish-German & 16,546 & + & + & 24.25 \\
    \end{tabular}
    \end{adjustbox}
    \caption{Available languages in the training set. Each language has its own annotation guidelines, in which capitalization can be taken into account (Caps), or words can be split or merged (1-N/N-1). Moreover, some languages are code-switched, two different languages are used in a tweet.}
    \label{tab:data}
\end{table}

\begin{table*}[hbt!]
    \centering
    \begin{adjustbox}{max width=\textwidth}
    \begin{tabular}{l|rr} 
    \textbf{Language} & \textbf{Example raw} & \textbf{Example gold}\\
    \hline\hline
    Croatian \cite{11356/1170} & dok je \textbf{bandic} bio \textbf{clan} \textbf{sdpa} tvrdilo se da je idealan & dok je \textbf{bandić} bio \textbf{član} \textbf{sdp-a} tvrdilo se da je idealan\\ 
    Danish \cite{plank-etal-2020-dan} & \textbf{Maerkeligt}, \textbf{taenker} jeg, og \textbf{gar} ind igen. & \textbf{Mærkeligt}, \textbf{tænker} jeg, og \textbf{går} ind igen. \\
    Dutch \cite{dutchNorm}  & \textbf{ja} \textbf{effe} slaapverhaal \textbf{vertelle} \textbf{vo} sophieke eh lol & \textbf{Ja} \textbf{even} slaapverhaal \textbf{vertellen} \textbf{voor} sophieke eh lol \\ 
    English \cite{baldwin-etal-2015-shared}  & he \textbf{obvi} \textbf{doesnt} understand that & he \textbf{obviously} \textbf{doesn't} understand that\\ 
    German \cite{sidarenka2013rule} & Ich \textbf{werd} \textbf{dran} denken! & Ich \textbf{werde} \textbf{daran} denken! \\ 
    Indonesian-English \cite{barik-etal-2019-normalization}  & \textbf{msh} \textbf{bs} disebut sukses? & \textbf{masih} \textbf{bisa} disebut sukses?\\ 
    Italian \cite{van-der-goot-etal-2020-norm} & \textbf{ztate} prentento in \textbf{ciro} \textbf{kvelli} quelli kol \textbf{raffrettoren} &  \textbf{state} prendendo in \textbf{giro} \textbf{quelli} col \textbf{raffreddore} \\ 
    Serbian \cite{11356/1171} & ja sam ozbiljan \textbf{covek} & ja sam ozbiljan \textbf{čovek} \\ 
    Slovenian  \cite{11356/1123} & da se \textbf{naujo} zdaj še na planico \textbf{spravl}!? & da se \textbf{ne} zdaj še na planico \textbf{spravili}!? \\ 
    Spanish \cite{alegria2013introduccion} & quiero \textbf{tranquileo} del bueno hoy..!!! & quiero \textbf{tranquilidad} del bueno hoy..!!! \\ 
    Turkish \cite{colakoglu-etal-2019-normalizing} & Avrupa ve \textbf{amerikada} \textbf{VALENTİNA} \textbf{DAY} diye geçer. & Avrupa ve \textbf{Amerika'da} \textbf{Valentina} \textbf{Day} diye geçer. \\ 
    Turkish-German \cite{van-der-goot-cetinoglu-2021-lexical} & \textbf{artik} ablamdan \textbf{bise} \textbf{yuruturum} \textbf{napim} :D & \textbf{Artık} ablamdan \textbf{bir şey} \textbf{yürütürüm} \textbf{ne yapayım} :D \\
    \end{tabular}
    \end{adjustbox}
    \caption{Noisy examples from each language and the corresponding canonical forms.}
    \label{tab:examples}
\end{table*}

\noindent\textbf{W-NUT Evaluation Methodology}
The organizers of the W-NUT workshop propose two types of evaluation procedures: \textit{intrinsic}, word-level and \textit{extrinsic}, downstream task performance (i.e. dependency parsing). 

As \textit{intrinsic} evaluation, the Error Reduction Rate introduced by \citet{van2019monoise} is proposed:

\begin{equation}
ERR = \frac{TP-FP}{TP+FN}
    \label{eqn:err}
\end{equation}

Because accuracy is hard to compare across datasets with different numbers of raw words which have to be normalized, the ERR proposes an evaluation metric that can be used to compare the performance of systems across multiple datasets. It is computed as accuracy normalized for the number of raw words normalized in the gold dataset.

A system that always keeps the raw words has an ERR score of 0.0, while a perfect system will have ERR precisely 1.0. The ERR has a negative value when the system normalizes more words with a wrong form than the correct canonical form.

However, one downside of the ERR is that it fails to distinguish between FP and FN. Thus, in the case of FP, the system may provide a correct normalization, even if the annotators did not normalize the raw word.

Further, two baselines are provided: \textbf{Leave-As-Is (LAI)} - the output is the same as the raw input, the normalization is not performed - and \textbf{Most-frequent-Replacement (MFR)} - the output is the most frequent replacement from the training data. If the raw word is not found in the training set, no normalization is performed.

As a secondary evaluation, the organizers propose an \textit{extrinsic} evaluation of the effect of normalization on the task of dependency parsing, previous research showing that lexical normalization improves the performance for this task \cite{van2019depth}. A dependency parser is trained on both raw and canonical data to evaluate the performance improvement of using the normalized versus the original data. 

Moreover, we also evaluate the extrinsic performance of our model on two additional tasks: sentiment analysis on the SMILE dataset \cite{wang2016smiles} and hate speech detection on OLID dataset \cite{zampieri-etal-2019-predicting}. Both datasets contain data collected from Twitter, making them good candidates for evaluating the semantic processing of noisy text.

\noindent \textbf{SMILE dataset} It consists of posts with mentions of several British museums gathered from Twitter to classify the emotions expressed by users towards art and cultural experiences from the museums. It contains 3,085 posts annotated with five emotions: anger, disgust, happiness, surprise and sadness; fear was not found in any Twitter posts.

\noindent \textbf{OLID dataset} It was the official dataset of the SemEval-2019 Task 6: Identifying and Categorizing Offensive Language in Social Media (OffensEval 2019) \cite{offenseval} and SemEval-2020 Task 12: Multilingual Offensive Language Identification in Social Media (OffensEval 2020) \cite{zampieri-etal-2020-semeval}. The dataset was also used in misogyny \cite{pamungkas2020misogyny}, cyberbullying \cite{aind2020q} and depression \cite{bucur-etal-2021-exploratory} research.
It contains 14,100 tweets with a hierarchical annotation taxonomy with three levels: Level A - Offensive language identification (offensive vs non-offensive), Level B - categorization of Offensive language (targeted insults or threats vs untargeted profanity) and Level C - Offensive language target identification (individual vs group vs other). However, for our evaluation, we focus only on level A. 

For evaluating on sentiment analysis (SMILE) and offensive language identification (OLID), we trained a simple word-level TF-IDF model together with a linear SVM with balanced weights. For SMILE, we report average macro F$_1$ score across 5 folds, and for OLID, we report macro F$_1$ score on the test set.

%% file: sections/4.Method.tex
\begin{figure}[hbt!]
    \centering
    \includegraphics[width=\linewidth]{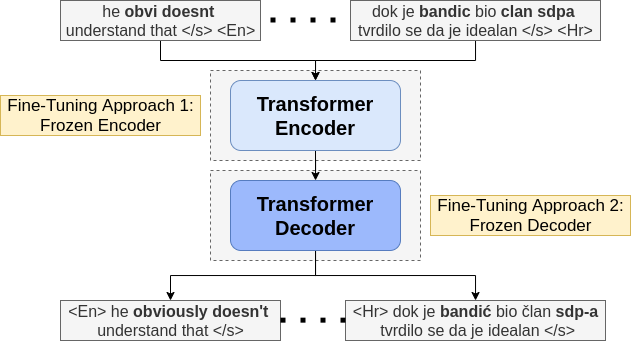}
    \caption{Fine-tuning a mBART model for lexical normalization on all available languages. We use the same model for all languages simultaneously. }
    \label{fig:fine-tuning}
\end{figure}

\input{tables/intrinsic}

\citet{lewis2019bart} proposed BART in 2019, as a way to pre-train large-scale transformers for sequence-to-sequence tasks. Initially, the authors pre-trained an encoder-decoder transformer only for English, obtaining good results on multiple downstream NLP tasks. Further, mBART \cite{tang2020multilingual}, follows the same procedure, but for multiple languages. The pre-training stage for both BART and mBART is akin to a denoising autoencoder, in which the model receives a noisy (in this case masked) sentence, and it learns to reconstruct it. 

While mBART is fine-tuned on multiple language pairs, it is pretrained monolingually, and is capable of acting as an autoencoder for the same language. In our case, we make use of a pretrained mBART on 50 languages\footnote{\url{https://huggingface.co/facebook/mbart-large-50}} from the transformers library \cite{wolf-etal-2020-transformers}, and employ a procedure similar to the pre-training stage: a noisy sentence is fed to the model, and the output ought to be the normalized sentence. We only trained on the provided languages that are contained in the pre-training set of languages. As mBART is not pre-trained on code-switched languages, for IN-EN and TR-DE, we use the mBART model pre-trained only on the main language of each pair (e.g. IN for IN-EN and TR for TR-DE).
For Danish and Serbian, we fall back to the MFR baseline.

Figure \ref{fig:fine-tuning} showcases our fine-tuning procedure. We tried two different approaches for fine-tuning: Frozen Encoder and Frozen Decoder because, with a fixed encoder, the model suffers from the same OOV-type problems as a typical transformer. However, training with a fixed decoder allows the model to better adapt its representations to each language's noisy version while maintaining its generative properties. For both approaches, we train a single model for all languages. Moreover, we also trained a separate mBART for each language, monolingually. 

\noindent\textbf{Training details} 
For all runs, we fine-tune mBART for 50 epochs, using a batch size of 256 and with a cyclical learning rate scheduler \cite{DBLP:journals/corr/Smith15a} that linearly increases the learning rate from 0.00001 to 0.0001 and back across 5 epochs. The workshop organizers provided both the training data and the validation data on most languages. We omit validation on languages where the validation data is missing. The training was performed on an NVIDIA RTX 2070 graphics card. Since the memory requirements of an mBART model are quite high, we employed gradient accumulation to increase the batch size. In addition, we employed early stopping when the validation loss increased for more than 3 epochs.

\noindent\textbf{Post-processing} Since our model outputs a whole sentence directly, the word-level evaluation requires the noisy input words to be aligned to their normalized counterpart. This phase is essential for sequence-to-sequence text normalization, as bad alignments will reduce the overall word-level performance score, especially in the 1-N/N-1 languages. As such, for the post-processing phase, we aligned input words with their normalized counterparts based on the Levenshtein distance between them. We used a linear sum assignment on the distance matrix to perform matching. Additionally, we matched the capitalization between corrections and left links, hashtags, and user mentions as they are.

%% file: tables/intrinsic.tex
\begin{table*}[hbt!]
    \centering
    \begin{adjustbox}{max width=\textwidth}
    \begin{tabular}{l|rrrrrrrrrrrrr} 
    \textbf{Team Name} & \textbf{Avg.} & \textbf{da} & \textbf{de} & \textbf{en} & \textbf{es} & \textbf{hr} & \textbf{iden} & \textbf{it} & \textbf{nl} & \textbf{sl} & \textbf{sr} & \textbf{tr} & \textbf{trde} \\ 
    \hline\hline
    ÚFAL-2 \cite{wnut-ufal} & \textbf{67.30} & 68.67 & \textbf{66.22} & \textbf{75.60} & 59.25 & \textbf{67.74} & \textbf{67.18} & \textbf{47.52} & \textbf{63.58} & \textbf{80.07} & \textbf{74.59} & \textbf{68.58} & \textbf{68.62} \\
    HEL-LJU-2 \cite{wnut-sinai} & 53.58 & 56.65 & 59.80 & 62.05 & 35.55 & 56.24 & 55.33 & 35.64 & 45.88 & 66.97 & 66.44 & 51.18 & 51.18 \\ 
    MoNoise \cite{van2019monoise} & 49.02 & 51.27 & 46.96 & 74.35 & 45.53 & 52.63 & 59.79 & 21.78 & 49.53 & 61.91 & 59.58 & 28.21 & 36.72 \\ 
    TrinkaAI-2$^*$ \cite{wnut-seqlab} & 43.75 & 45.89 & 47.30 & 65.96 & \textbf{61.33} & 41.28 & 56.36 & 15.84 & 45.74 & 59.51 & 44.52 & 15.54 & 25.77 \\ 
    thunderml-1$^*$ & 43.44 & 46.52 & 46.62 & 64.07 & 60.29 & 40.09 & 59.11 & 11.88 & 44.05 & 59.33 & 44.46 & 15.88 & 29.01 \\ 
    team-2 & 40.70 & 48.10 & 46.06 & 63.73 & 21.00 & 40.39 & 59.28 & 13.86 & 43.72 & 60.55 & 46.11 & 15.88 & 29.71 \\ 
    learnML-2 & 40.30 & 40.51 & 43.69 & 61.57 & 56.55 & 38.11 & 56.19 & 5.94 & 42.77 & 58.25 & 39.99 & 14.36 & 25.68 \\ 
    maet-1 & 40.05 & 48.10 & 46.06 & 63.90 & 21.00 & 40.39 & 59.28 & 5.94 & 43.72 & 60.55 & 46.11 & 15.88 & 29.71 \\
    MFR & 38.37 & 49.68 & 32.09 & 64.93 & 25.57 & 36.52 & 61.17 & 16.83 & 37.70 & 56.71 & 42.62 & 14.53 & 22.09 \\
    CL-MoNoise \cite{wnut-clmonoise} & 12.05 & 7.28 & 16.55 & 4.13 & 4.99 & 26.41 & 2.41 & 0.00 & 16.22 & 8.77 & 20.09 & 17.57 & 20.16 \\
    \hline
    (ours) Fixed Encoder (separate) + post proc.    & 10.65 & \textit{49.68}  & -2.59 & 29.13 & -7.90 & 26.41     & -1.72  & -8.91      & -1.49 & 1.27  & \textit{42.62} & 0.68 & 0.70 \\
    (ours) Fixed Encoder (separate)                 & 6.73  & \textit{49.68} & -1.91 & 26.81 & -9.36 & -10.06    & -7.22 & -8.91      & -2.09 & -1.04 & \textit{42.62} & 9.97  & 14.99 \\
    (ours) Fixed Encoder (separate) - stripped unicode  & 5.22  & \textit{49.68} & -1.91 & 26.81 & -10.19 & -9.86    & -7.22 & -31.68     & -2.09 & -1.13 & \textit{42.62} & 1.01 & 6.57 \\
    (ours) ML - Fixed Decoder + post proc.          & -6.54 & \textit{49.68} & 12.50 & 27.41 & -13.10 & -111.84 & -7.73     & -8.91     & 16.82      & -110.57 & \textit{42.62} & 11.49 & 13.06\\
    (ours) ML - Fixed Decoder                       & -11.79 & \textit{49.68} & 20.05 & 22.12 & -18.92 & -127.60 & -14.60     & -25.74    & 16.69     & -133.71 & \textit{42.62} & 13.18 & 14.64 \\
    
    (ours) ML - Fixed Encoder + post proc.          & -21.51 & \textit{49.68} & 10.47 & 12.09 & -28.69 & -191.33 & -9.97 & -27.72    & 9.19  & -141.80 & \textit{42.62}     & 9.63 & 7.62 \\
    (ours) ML - Fixed Encoder                       & -32.90 & \textit{49.68} & 19.48 & 20.78 & -40.12 & -242.57 & -24.23 & -70.30   & 8.72  & -180.75 & \textit{42.62}     & 11.15 & 10.69 \\
    
    \hline
    LAI & 0.00 & 0.00 & 0.00 & 0.00 & 0.00 & 0.00 & 0.00 & 0.00 & 0.00 & 0.00 & 0.00 & 0.00 & 0.00 \\
    MaChAmp \cite{van-der-goot-etal-2021-massive} & -21.25 & -88.92 & -93.36 & 50.99 & 25.36 & 42.62 & 39.52 & -312.87 & 1.49 & 56.80 & 39.44 & -12.67 & -3.42
    \end{tabular}
    \end{adjustbox}
    \caption{Team standings, based on Error Reduction Rate (ERR). We kept the best result from each team, from clarity. ($^*$ denotes late submissions).}
    \label{tab:intrinsic-evaluation}
\end{table*}

%% file: sections/5.Results.tex
\input{tables/extrinsic}

\begin{table}[hbt!]
    \centering
    \begin{adjustbox}{max width=\linewidth}
    \begin{tabular}{l|cc}
         & SMILE & OLID Task A \\
        \hline\hline
        Raw Text (Leave-As-Is) & 22.65\% $\pm$ 0.02 & 57.15\%\\
        mBART Fixed Encoder & \textbf{ 23.43\% $\pm$ 0.02} & \textbf{58.08\%}\\

    \end{tabular}
    \end{adjustbox}
    \caption{Extrinsic evaluation on sentiment analysis (SMILE) and offensive language identification (OLID). Lexical normalization through fine-tuning mBART slightly improves performance.}
    \label{tab:our-extrinsic}
\end{table}

\begin{table}[hbt!]
\renewcommand{\arraystretch}{1.2}
    \centering
    \resizebox{\linewidth}{!}{
        \begin{tabular}{p{2.5cm}p{2.5cm}p{2.5cm}|c}
            \textbf{Raw} & \textbf{Gold} & \textbf{Our} & \textbf{Correct?}\\
            \hline\hline
            i see, u can comeee & i see, you can come & i see, you can come & \checkmark\\
            \hline
            ich geb heut einen aus & ich gebe heute einen aus & ich gebe heute einen aus & \checkmark\\
            \hline
            Juhuuuuu & Juhu  & Juhu & \checkmark \\
            \hline\hline
            fakt ap gaan. eig nu al mr kanniet & fakt op gaan. Eigenlijk nu al maar kan niet & \textbf{echt} ap gaan. eig nu al \textbf{mijn kanniet} & \xmark \\
            \hline
             "Why Germany says "nein" &  "Why Germany says "nein"& "\textbf{Warum Deutschland sagt} "nein" & \xmark \\
             \hline
             i coulda swore .... lol nvm & i could have swore .... lol never mind & i \textbf{could} swore .... lol \textbf{never} & \xmark \\ 
             \hline\hline
             todos lo sabemos \textbf{jajajajajaja} & todos lo sabemos \textbf{jajajajajaja} & todos lo sabemos \textbf{ja} & \qm \\
             \hline
             discussing \textbf{w/} friend & discussing \textbf{w/} friend & discussing \textbf{with} friend & \qm\\
            \hline
            \textbf{n} puedo ni volver a dormirme & \textbf{n} puedo ni volver a dormirme & \textbf{no} puedo ni volver a dormirme & \qm \\
        \end{tabular}
    }
    \caption{Qualitative results on different languages with mBART Fixed Encoder. We present examples of correct normalization (\checkmark), mistakes (\xmark), and questionable normalizations (\qm), in which the model correctly normalizes, but annotators do not.}
    \label{tab:qualitative}
\end{table}

We further showcase the results of the pretrained mBART models fine-tuned on the available data: firstly, we kept the transformer encoder fixed and trained only the decoder, and secondly, we kept the decoder fixed and trained the encoder. During this fine-tuning process, we trained a single model for all languages. Further, for the CodaLab submission, we fine-tuned multiple models, one for each language, in the "fixed encoder" regime. 

\noindent\textbf{Intrinsic Evaluation} Table \ref{tab:intrinsic-evaluation} showcases intrinsic, word-level evaluation across languages. Our best model obtained an average ERR across languages of 10.65, corresponding to a separate mBART trained for each language, with the additional post-processing described in Section 4. In our case, training multilingually did improve performance on some languages (i.e. \textit{DE, EN, NL, TR}), but overall achieved lower scores, especially in the case for \textit{HR} and \textit{SL}. For those languages, the model severely diverged, and its output consisted only of repeating the first word in the sentence. As per our intuition, fixing the decoder results in better performance when compared to fixing the encoder: the model learns to adapt its representations to account for the noisy text. 

However, since our method is sentence-based, perfect alignment between words is cumbersome, with many cases of mismatch between punctuation. Moreover, merging or splitting words for normalization is also not taken into account in the post-processing phase.

\noindent\textbf{Extrinsic Evaluation} For extrinsic evaluation, we showcase the results for our best model in Table \ref{tab:extrinsic-evaluation} on the dependency parsing downstream task from the workshop challenge. Even though our model is not in the top-performing models, the absolute difference in performance is minimal.

Moreover, we also evaluated the effect of lexical normalization on two other tasks - sentiment analysis on the SMILE dataset and offensive language identification on OLID (Table \ref{tab:our-extrinsic}). We trained a word-level TF-IDF and a linear SVM with balanced weights for both datasets and reported a macro F$_1$ score.  Our lexical normalization improves results on both these tasks, compared to modelling the raw, unprocessed social media posts. This is because lexical normalization results in a smaller vocabulary for the documents, allowing the SVM model to operate on smaller dimensional data. Moreover, this evaluation procedure is arguably more realistic, as it does not require accurate post-processing to precisely align noisy words with their corrected version and match punctuation.

\noindent\textbf{Discussion} In Table \ref{tab:qualitative} we showcase some examples for correct, incorrect and questionable text normalizations. The model is able to easily grasp contractions such as \textit{u} $\rightarrow$ \textit{you} and expressive lengthening such as \textit{Juhuuuuu} $\rightarrow$ \textit{Juhu}. However, more complex word abbreviations such as \textit{nvm} are quite challenging to generate, as the model only outputs the first part (i.e. \textit{never}). Moreover, code-switched languages are an inherent problem to our approach, as mBART is only trained to receive input from a single language and not code-switched. Interestingly, even though we specified the langauge code for German in the phrase \textit{"Why Germany says "nein"}, the model actually translates the English part into German: \textit{Warum Deutschland sagt "nein"}.

However, as the organizers have pointed out, there are inconsistencies in the training and testing data annotations. In some cases, some words are not normalized (i.e. \textit{jajajajaj} / \textit{w/} / \textit{n}) even though they were clearly lengthenings or contractions. Despite this, in some of these cases, our model was able to provide correct normalizations.

There also appears to be no correlation between training dataset size and final normalization performance. For example, in the case of Croatian, even though the dataset is the second largest, the performance is lower than for other languages. Thus, the lower performance in some languages may be a cause of the complexity of the language; for English, our model obtained the best results.

%% file: tables/extrinsic.tex
\begin{table*}[hbt!]
    \centering
    \begin{adjustbox}{max width=\textwidth}
    \begin{tabular}{lrrrrrrrr} 
    \textbf{Team Name} & \textbf{Avg.} & \textbf{de-tweede} & \textbf{en-aae} & \textbf{en-monoise} & \textbf{en-tweebank2} & \textbf{it-postwita} & \textbf{it-twittiro} & \textbf{tr-iwt151} \\
    \hline\hline
    ÚFAL-2 & \textbf{64.17} & \textbf{73.58} & \textbf{62.73} & \textbf{58.57} & 59.08 & \textbf{68.28} & 72.22 & 54.74 \\
    HEL-LJU-2 & 63.73 & 73.49 & 60.64 & 56.27 & 60.30 & 68.11 & 72.32 & 54.95 \\
    MoNoise & 63.44 & 73.20 & 62.27 & 56.83 & 58.90 & 67.55 & 70.69 & 54.61 \\
    MFR & 63.31 & 72.86 & 60.32 & 56.74 & \textbf{60.31} & 67.34 & 70.72 & 54.89 \\
    TrinkaAI-2* & 63.12 & 72.86 & 60.16 & 56.64 & 59.87 & 66.98 & 71.14 & 54.20 \\
    maet-1 & 63.09 & 72.80 & 59.44 & 56.64 & 59.80 & 67.41 & 71.07 & 54.45 \\
    team-2 & 63.03 & 72.80 & 59.44 & 56.64 & 59.80 & 67.19 & 70.86 & 54.45 \\
    thunderml-2* & 63.02 & 72.67 & 59.57 & 56.74 & 59.25 & 67.34 & 71.35 & 54.24 \\
    thunderml-1* & 62.95 & 72.52 & 59.31 & 56.74 & 59.86 & 67.09 & 71.00 & 54.09 \\
    CL-MoNoise & 62.71 & 72.65 & 60.90 & 55.26 & 58.53 & 66.53 & 70.10 & 54.98 \\
    \hline
    (ours) Fixed Encoder (separate) & 62.53 & 72.57 & 59.57 & 54.20 & 59.81 & 66.74 & 69.99 & 54.84 \\
    \hline
    LAI & 62.45 & 72.71 & 59.21 & 53.65 & 59.99 & 66.49 & 70.06 & \textbf{55.00}\\
    MaChAmp & 61.89 & 71.28 & 60.77 & 54.61 & 57.97 & 64.65 & 69.82 & 54.08
    \end{tabular}
    \end{adjustbox}
    \caption{Extrinsic evaluation results on dependency parsing task.}
    \label{tab:extrinsic-evaluation}
\end{table*}

%% file: emnlp2021.bbl
\begin{thebibliography}{61}
\expandafter\ifx\csname natexlab\endcsname\relax\def\natexlab#1{#1}\fi

\bibitem[{Aind et~al.(2020)Aind, Ramnaney, and Sethia}]{aind2020q}
Alwin~T Aind, Akashdeep Ramnaney, and Divyashikha Sethia. 2020.
\newblock Q-bully: a reinforcement learning based cyberbullying detection
  framework.
\newblock In \emph{2020 International Conference for Emerging Technology
  (INCET)}, pages 1--6. IEEE.

\bibitem[{Akhtar et~al.(2015)Akhtar, Sikdar, and Ekbal}]{akhtar2015iitp}
Md~Shad Akhtar, Utpal~Kumar Sikdar, and Asif Ekbal. 2015.
\newblock Iitp: Hybrid approach for text normalization in twitter.
\newblock In \emph{Proceedings of the Workshop on Noisy User-generated Text},
  pages 106--110.

\bibitem[{Alegria et~al.(2013)Alegria, Aranberri, Fresno, Gamallo, Padr{\'o},
  San~Vicente, Turmo, and Zubiaga}]{alegria2013introduccion}
Inaki Alegria, Nora Aranberri, V{\'\i}ctor Fresno, Pablo Gamallo, Lluis
  Padr{\'o}, Inaki San~Vicente, Jordi Turmo, and Arkaitz Zubiaga. 2013.
\newblock Introducci{\'o}n a la tarea compartida {Tweet-Norm} 2013:
  Normalizaci{\'o}n l{\'e}xica de tuits en {E}spa{\~n}ol.
\newblock In \emph{Tweet-Norm@ SEPLN}, pages 1--9.

\bibitem[{Amir et~al.(2019)Amir, Dredze, and Ayers}]{amir2019mental}
Silvio Amir, Mark Dredze, and John~W Ayers. 2019.
\newblock Mental health surveillance over social media with digital cohorts.
\newblock In \emph{Proceedings of the Sixth Workshop on Computational
  Linguistics and Clinical Psychology}, pages 114--120.

\bibitem[{Baldwin et~al.(2015{\natexlab{a}})Baldwin, de~Marneffe, Han, Kim,
  Ritter, and Xu}]{baldwin2015shared}
Timothy Baldwin, Marie-Catherine de~Marneffe, Bo~Han, Young-Bum Kim, Alan
  Ritter, and Wei Xu. 2015{\natexlab{a}}.
\newblock Shared tasks of the 2015 workshop on noisy user-generated text:
  Twitter lexical normalization and named entity recognition.
\newblock In \emph{Proceedings of the Workshop on Noisy User-generated Text},
  pages 126--135.

\bibitem[{Baldwin et~al.(2015{\natexlab{b}})Baldwin, de~Marneffe, Han, Kim,
  Ritter, and Xu}]{baldwin-etal-2015-shared}
Timothy Baldwin, Marie~Catherine de~Marneffe, Bo~Han, Young-Bum Kim, Alan
  Ritter, and Wei Xu. 2015{\natexlab{b}}.
\newblock \href {https://doi.org/10.18653/v1/W15-4319} {Shared tasks of the
  2015 workshop on noisy user-generated text: {T}witter lexical normalization
  and named entity recognition}.
\newblock In \emph{Proceedings of the Workshop on Noisy User-generated Text},
  pages 126--135, Beijing, China. Association for Computational Linguistics.

\bibitem[{Barik et~al.(2019)Barik, Mahendra, and
  Adriani}]{barik-etal-2019-normalization}
Anab~Maulana Barik, Rahmad Mahendra, and Mirna Adriani. 2019.
\newblock \href {https://doi.org/10.18653/v1/D19-5554} {Normalization of
  {I}ndonesian-{E}nglish code-mixed {T}witter data}.
\newblock In \emph{Proceedings of the 5th Workshop on Noisy User-generated Text
  (W-NUT 2019)}, pages 417--424, Hong Kong, China. Association for
  Computational Linguistics.

\bibitem[{Bucur et~al.(2021{\natexlab{a}})Bucur, Cosma, and
  Dinu}]{bucur2021early}
Ana-Maria Bucur, Adrian Cosma, and Liviu~P Dinu. 2021{\natexlab{a}}.
\newblock Early risk detection of pathological gambling, self-harm and
  depression using bert.
\newblock \emph{CLEF (Working Notes)}.

\bibitem[{Bucur et~al.(2021{\natexlab{b}})Bucur, Zampieri, and
  Dinu}]{bucur-etal-2021-exploratory}
Ana-Maria Bucur, Marcos Zampieri, and Liviu~P. Dinu. 2021{\natexlab{b}}.
\newblock \href {https://doi.org/10.18653/v1/2021.findings-acl.315} {An
  exploratory analysis of the relation between offensive language and mental
  health}.
\newblock In \emph{Findings of the Association for Computational Linguistics:
  ACL-IJCNLP 2021}, pages 3600--3606, Online. Association for Computational
  Linguistics.

\bibitem[{Cao et~al.(2019)Cao, Zhang, Feng, Wei, Wang, Li, and
  He}]{cao2019latent}
Lei Cao, Huijun Zhang, Ling Feng, Zihan Wei, Xin Wang, Ningyun Li, and Xiaohao
  He. 2019.
\newblock Latent suicide risk detection on microblog via suicide-oriented word
  embeddings and layered attention.
\newblock In \emph{Proceedings of the 2019 Conference on Empirical Methods in
  Natural Language Processing and the 9th International Joint Conference on
  Natural Language Processing (EMNLP-IJCNLP)}, pages 1718--1728.

\bibitem[{{\c{C}}olako{\u{g}}lu et~al.(2019){\c{C}}olako{\u{g}}lu, Sulubacak,
  and Tantu{\u{g}}}]{colakoglu-etal-2019-normalizing}
Talha {\c{C}}olako{\u{g}}lu, Umut Sulubacak, and Ahmet~C{\"u}neyd Tantu{\u{g}}.
  2019.
\newblock \href {https://doi.org/10.18653/v1/P19-2037} {Normalizing
  non-canonical {T}urkish texts using machine translation approaches}.
\newblock In \emph{Proceedings of the 57th Annual Meeting of the Association
  for Computational Linguistics: Student Research Workshop}, pages 267--272,
  Florence, Italy. Association for Computational Linguistics.

\bibitem[{Coppersmith et~al.(2014)Coppersmith, Harman, and
  Dredze}]{coppersmith2014measuring}
Glen Coppersmith, Craig Harman, and Mark Dredze. 2014.
\newblock Measuring post traumatic stress disorder in twitter.
\newblock In \emph{Eighth international AAAI conference on weblogs and social
  media}.

\bibitem[{Dekker and van~der Goot(2020)}]{dekker2020synthetic}
Kelly Dekker and Rob van~der Goot. 2020.
\newblock Synthetic data for english lexical normalization: How close can we
  get to manually annotated data?
\newblock In \emph{Proceedings of the 12th Language Resources and Evaluation
  Conference}, pages 6300--6309.

\bibitem[{Eisenstein(2013)}]{eisenstein2013bad}
Jacob Eisenstein. 2013.
\newblock What to do about bad language on the internet.
\newblock In \emph{Proceedings of the 2013 conference of the North American
  Chapter of the association for computational linguistics: Human language
  technologies}, pages 359--369.

\bibitem[{Erjavec et~al.(2017)Erjavec, Fi{\v s}er, {\v C}ibej, Arhar~Holdt,
  Ljube{\v s}i{\'c}, and Zupan}]{11356/1123}
Toma{\v z} Erjavec, Darja Fi{\v s}er, Jaka {\v C}ibej, {\v S}pela Arhar~Holdt,
  Nikola Ljube{\v s}i{\'c}, and Katja Zupan. 2017.
\newblock \href {http://hdl.handle.net/11356/1123} {{CMC} training corpus
  {Janes-Tag} 2.0}.
\newblock Slovenian language resource repository {CLARIN}.{SI}.

\bibitem[{Guntuku et~al.(2019)Guntuku, Schneider, Pelullo, Young, Wong, Ungar,
  Polsky, Volpp, and Merchant}]{guntuku2019studying}
Sharath~Chandra Guntuku, Rachelle Schneider, Arthur Pelullo, Jami Young, Vivien
  Wong, Lyle Ungar, Daniel Polsky, Kevin~G Volpp, and Raina Merchant. 2019.
\newblock Studying expressions of loneliness in individuals using twitter: an
  observational study.
\newblock \emph{BMJ open}, 9(11):e030355.

\bibitem[{Jin(2015)}]{jin2015ncsu}
Ning Jin. 2015.
\newblock Ncsu-sas-ning: Candidate generation and feature engineering for
  supervised lexical normalization.
\newblock In \emph{Proceedings of the Workshop on Noisy User-generated Text},
  pages 87--92.

\bibitem[{Kubal and Nagvenkar(2021)}]{wnut-seqlab}
Divesh Kubal and Apurva Nagvenkar. 2021.
\newblock Multilingual sequence labeling approach to solve lexical
  normalization.
\newblock In \emph{Proceedings of the 7th Workshop on Noisy User-generated Text
  (W-NUT 2021)}, Punta Cana, Dominican Republic. Association for Computational
  Linguistics.

\bibitem[{Kumar et~al.(2020)Kumar, Makhija, and Gupta}]{kumar-etal-2020-noisy}
Ankit Kumar, Piyush Makhija, and Anuj Gupta. 2020.
\newblock \href {https://doi.org/10.18653/v1/2020.wnut-1.3} {Noisy text data:
  Achilles{'} heel of {BERT}}.
\newblock In \emph{Proceedings of the Sixth Workshop on Noisy User-generated
  Text (W-NUT 2020)}, pages 16--21, Online. Association for Computational
  Linguistics.

\bibitem[{Lewis et~al.(2019)Lewis, Liu, Goyal, Ghazvininejad, Mohamed, Levy,
  Stoyanov, and Zettlemoyer}]{lewis2019bart}
Mike Lewis, Yinhan Liu, Naman Goyal, Marjan Ghazvininejad, Abdelrahman Mohamed,
  Omer Levy, Ves Stoyanov, and Luke Zettlemoyer. 2019.
\newblock Bart: Denoising sequence-to-sequence pre-training for natural
  language generation, translation, and comprehension.
\newblock \emph{arXiv preprint arXiv:1910.13461}.

\bibitem[{Ljube{\v{s}}i{\'c} et~al.(2014)Ljube{\v{s}}i{\'c}, Erjavec, and
  Fi{\v{s}}er}]{ljubevsic2014standardizing}
Nikola Ljube{\v{s}}i{\'c}, Toma{\v{z}} Erjavec, and Darja Fi{\v{s}}er. 2014.
\newblock Standardizing tweets with character-level machine translation.
\newblock In \emph{International Conference on Intelligent Text Processing and
  Computational Linguistics}, pages 164--175. Springer.

\bibitem[{Ljube{\v s}i{\'c} et~al.(2017{\natexlab{a}})Ljube{\v s}i{\'c},
  Erjavec, Mili{\v c}evi{\'c}, and Samard{\v z}i{\'c}}]{11356/1170}
Nikola Ljube{\v s}i{\'c}, Toma{\v z} Erjavec, Maja Mili{\v c}evi{\'c}, and
  Tanja Samard{\v z}i{\'c}. 2017{\natexlab{a}}.
\newblock \href {http://hdl.handle.net/11356/1170} {Croatian {T}witter training
  corpus {ReLDI}-{NormTagNER}-hr 2.0}.
\newblock Slovenian language resource repository {CLARIN}.{SI}.

\bibitem[{Ljube{\v s}i{\'c} et~al.(2017{\natexlab{b}})Ljube{\v s}i{\'c},
  Erjavec, Mili{\v c}evi{\'c}, and Samard{\v z}i{\'c}}]{11356/1171}
Nikola Ljube{\v s}i{\'c}, Toma{\v z} Erjavec, Maja Mili{\v c}evi{\'c}, and
  Tanja Samard{\v z}i{\'c}. 2017{\natexlab{b}}.
\newblock \href {http://hdl.handle.net/11356/1171} {Serbian {T}witter training
  corpus {ReLDI}-{NormTagNER}-sr 2.0}.
\newblock Slovenian language resource repository {CLARIN}.{SI}.

\bibitem[{Lourentzou et~al.(2019)Lourentzou, Manghnani, and
  Zhai}]{lourentzou2019adapting}
Ismini Lourentzou, Kabir Manghnani, and ChengXiang Zhai. 2019.
\newblock Adapting sequence to sequence models for text normalization in social
  media.
\newblock In \emph{Proceedings of the International AAAI Conference on Web and
  Social Media}, volume~13, pages 335--345.

\bibitem[{Lusetti et~al.(2018)Lusetti, Ruzsics, G{\"o}hring, Samardzic, and
  Stark}]{lusetti2018encoder}
Massimo Lusetti, Tatyana Ruzsics, Anne G{\"o}hring, Tanja Samardzic, and
  Elisabeth Stark. 2018.
\newblock Encoder-decoder methods for text normalization.
\newblock In \emph{Proceedings of the Fifth Workshop on NLP for Similar
  Languages, Varieties and Dialects (VarDial 2018)}, pages 18--28.

\bibitem[{Ma(2019)}]{ma2019nlpaug}
Edward Ma. 2019.
\newblock Nlp augmentation.
\newblock https://github.com/makcedward/nlpaug.

\bibitem[{Mandal and Nanmaran(2018)}]{mandal2018normalization}
Soumil Mandal and Karthick Nanmaran. 2018.
\newblock Normalization of transliterated words in code-mixed data using
  seq2seq model \& levenshtein distance.
\newblock In \emph{Proceedings of the 2018 EMNLP Workshop W-NUT: The 4th
  Workshop on Noisy User-generated Text}, pages 49--53.

\bibitem[{Matero et~al.(2019)Matero, Idnani, Son, Giorgi, Vu, Zamani,
  Limbachiya, Guntuku, and Schwartz}]{matero2019suicide}
Matthew Matero, Akash Idnani, Youngseo Son, Salvatore Giorgi, Huy Vu, Mohammad
  Zamani, Parth Limbachiya, Sharath~Chandra Guntuku, and H~Andrew Schwartz.
  2019.
\newblock Suicide risk assessment with multi-level dual-context language and
  bert.
\newblock In \emph{Proceedings of the sixth workshop on computational
  linguistics and clinical psychology}, pages 39--44.

\bibitem[{Min and Mott(2015)}]{min2015ncsu_sas_wookhee}
Wookhee Min and Bradford Mott. 2015.
\newblock Ncsu\_sas\_wookhee: A deep contextual long-short term memory model
  for text normalization.
\newblock In \emph{Proceedings of the Workshop on Noisy User-generated Text},
  pages 111--119.

\bibitem[{Muller et~al.(2019)Muller, Sagot, and Seddah}]{muller2019enhancing}
Benjamin Muller, Beno{\^\i}t Sagot, and Djam{\'e} Seddah. 2019.
\newblock Enhancing bert for lexical normalization.
\newblock In \emph{The 5th Workshop on Noisy User-generated Text (W-NUT)}.

\bibitem[{Nguyen et~al.(2021)Nguyen, Rosseel, and Grieve}]{nguyen2021learning}
Dong Nguyen, Laura Rosseel, and Jack Grieve. 2021.
\newblock On learning and representing social meaning in nlp: a sociolinguistic
  perspective.
\newblock In \emph{Proceedings of the 2021 Conference of the North American
  Chapter of the Association for Computational Linguistics: Human Language
  Technologies}, pages 603--612.

\bibitem[{Pamungkas et~al.(2020)Pamungkas, Basile, and
  Patti}]{pamungkas2020misogyny}
Endang~Wahyu Pamungkas, Valerio Basile, and Viviana Patti. 2020.
\newblock Misogyny detection in twitter: a multilingual and cross-domain study.
\newblock \emph{Information Processing \& Management}, 57(6):102360.

\bibitem[{Pennell and Liu(2011)}]{pennell2011character}
Deana Pennell and Yang Liu. 2011.
\newblock A character-level machine translation approach for normalization of
  sms abbreviations.
\newblock In \emph{Proceedings of 5th International Joint Conference on Natural
  Language Processing}, pages 974--982.

\bibitem[{Plank et~al.(2020)Plank, Jensen, and van~der
  Goot}]{plank-etal-2020-dan}
Barbara Plank, Kristian~N{\o}rgaard Jensen, and Rob van~der Goot. 2020.
\newblock \href {https://doi.org/10.18653/v1/2020.coling-main.583} {{D}a{N}+:
  {D}anish nested named entities and lexical normalization}.
\newblock In \emph{Proceedings of the 28th International Conference on
  Computational Linguistics}, pages 6649--6662, Barcelona, Spain (Online).
  International Committee on Computational Linguistics.

\bibitem[{Samuel and Straka(2021)}]{wnut-ufal}
David Samuel and Milan Straka. 2021.
\newblock {ÚFAL} at {MultiLexNorm} 2021: Improving multilingual lexical
  normalization by fine-tuning {ByT5}.
\newblock In \emph{Proceedings of the 7th Workshop on Noisy User-generated Text
  (W-NUT 2021)}, Punta Cana, Dominican Republic. Association for Computational
  Linguistics.

\bibitem[{Scherrer and Ljubešić(2021)}]{wnut-sinai}
Yves Scherrer and Nikola Ljubešić. 2021.
\newblock Sesame {S}treet to {M}ount {S}inai: {BERT}-constrained
  character-level {M}oses models for multilingual lexical normalization.
\newblock In \emph{Proceedings of the 7th Workshop on Noisy User-generated Text
  (W-NUT 2021)}, Punta Cana, Dominican Republic. Association for Computational
  Linguistics.

\bibitem[{Schuur(2020)}]{dutchNorm}
Youri Schuur. 2020.
\newblock \href {http://robvanderg.github.io/doc/normalization_dutch.pdf}
  {Normalization for {D}utch for improved pos tagging}.
\newblock Master's thesis, University of Groningen.

\bibitem[{Sidarenka et~al.(2013)Sidarenka, Scheffler, and
  Stede}]{sidarenka2013rule}
Uladzimir Sidarenka, Tatjana Scheffler, and Manfred Stede. 2013.
\newblock Rule-based normalization of {G}erman {T}witter messages.
\newblock In \emph{Proc. of the GSCL Workshop Verarbeitung und Annotation von
  Sprachdaten aus Genres internetbasierter Kommunikation}.

\bibitem[{Smith(2015)}]{DBLP:journals/corr/Smith15a}
Leslie~N. Smith. 2015.
\newblock \href {http://arxiv.org/abs/1506.01186} {No more pesky learning rate
  guessing games}.
\newblock \emph{CoRR}, abs/1506.01186.

\bibitem[{Supranovich and Patsepnia(2015)}]{supranovich2015ihs_rd}
Dmitry Supranovich and Viachaslau Patsepnia. 2015.
\newblock Ihs\_rd: Lexical normalization for english tweets.
\newblock In \emph{Proceedings of the Workshop on Noisy User-generated Text},
  pages 78--81.

\bibitem[{Tadesse et~al.(2019)Tadesse, Lin, Xu, and
  Yang}]{tadesse2019detection}
Michael~M Tadesse, Hongfei Lin, Bo~Xu, and Liang Yang. 2019.
\newblock Detection of depression-related posts in reddit social media forum.
\newblock \emph{IEEE Access}, 7:44883--44893.

\bibitem[{Tang et~al.(2020)Tang, Tran, Li, Chen, Goyal, Chaudhary, Gu, and
  Fan}]{tang2020multilingual}
Yuqing Tang, Chau Tran, Xian Li, Peng-Jen Chen, Naman Goyal, Vishrav Chaudhary,
  Jiatao Gu, and Angela Fan. 2020.
\newblock Multilingual translation with extensible multilingual pretraining and
  finetuning.
\newblock \emph{arXiv preprint arXiv:2008.00401}.

\bibitem[{van~der Goot(2019{\natexlab{a}})}]{van2019depth}
Rob van~der Goot. 2019{\natexlab{a}}.
\newblock An in-depth analysis of the effect of lexical normalization on the
  dependency parsing of social media.
\newblock In \emph{Proceedings of the 5th Workshop on Noisy User-generated Text
  (W-NUT 2019)}, pages 115--120.

\bibitem[{van~der Goot(2019{\natexlab{b}})}]{van2019monoise}
Rob van~der Goot. 2019{\natexlab{b}}.
\newblock Monoise: A multi-lingual and easy-to-use lexical normalization tool.
\newblock In \emph{Proceedings of the 57th Annual Meeting of the Association
  for Computational Linguistics: System Demonstrations}, pages 201--206.

\bibitem[{van~der Goot(2021)}]{wnut-clmonoise}
Rob van~der Goot. 2021.
\newblock {CL-MoNoise}: Cross-lingual lexical normalization.
\newblock In \emph{Proceedings of the 7th Workshop on Noisy User-generated Text
  (W-NUT 2021)}, Punta Cana, Dominican Republic. Association for Computational
  Linguistics.

\bibitem[{van~der Goot and
  {\c{C}}etino{\u{g}}lu(2021)}]{van-der-goot-cetinoglu-2021-lexical}
Rob van~der Goot and {\"O}zlem {\c{C}}etino{\u{g}}lu. 2021.
\newblock \href {https://aclanthology.org/2021.eacl-main.200} {Lexical
  normalization for code-switched data and its effect on {POS} tagging}.
\newblock In \emph{Proceedings of the 16th Conference of the European Chapter
  of the Association for Computational Linguistics: Main Volume}, pages
  2352--2365, Online. Association for Computational Linguistics.

\bibitem[{van~der Goot et~al.(2020)van~der Goot, Ramponi, Caselli, Cafagna, and
  De~Mattei}]{van-der-goot-etal-2020-norm}
Rob van~der Goot, Alan Ramponi, Tommaso Caselli, Michele Cafagna, and Lorenzo
  De~Mattei. 2020.
\newblock \href {https://www.aclweb.org/anthology/2020.lrec-1.769} {Norm it!
  lexical normalization for {I}talian and its downstream effects for dependency
  parsing}.
\newblock In \emph{Proceedings of the 12th Language Resources and Evaluation
  Conference}, pages 6272--6278, Marseille, France. European Language Resources
  Association.

\bibitem[{van~der Goot et~al.(2021{\natexlab{a}})van~der Goot, Ramponi,
  Zubiaga, Plank, Muller, San Vicente~Roncal, Ljube{\v{s}}i\'{c},
  {\c{C}}etino{\u{g}}lu, Mahendra, {\c{C}}olako{\u{g}}lu, Baldwin, Caselli, and
  Sidorenko}]{multilexnorm}
Rob van~der Goot, Alan Ramponi, Arkaitz Zubiaga, Barbara Plank, Benjamin
  Muller, I\~{n}aki San Vicente~Roncal, Nikola Ljube{\v{s}}i\'{c}, {\"O}zlem
  {\c{C}}etino{\u{g}}lu, Rahmad Mahendra, Talha {\c{C}}olako{\u{g}}lu, Timothy
  Baldwin, Tommaso Caselli, and Wladimir Sidorenko. 2021{\natexlab{a}}.
\newblock {MultiLexNorm}: A shared task on multilingual lexical normalization.
\newblock In \emph{Proceedings of the 7th Workshop on Noisy User-generated Text
  (W-NUT 2021)}, Punta Cana, Dominican Republic. Association for Computational
  Linguistics.

\bibitem[{van~der Goot et~al.(2021{\natexlab{b}})van~der Goot, {\"U}st{\"u}n,
  Ramponi, Sharaf, and Plank}]{van-der-goot-etal-2021-massive}
Rob van~der Goot, Ahmet {\"U}st{\"u}n, Alan Ramponi, Ibrahim Sharaf, and
  Barbara Plank. 2021{\natexlab{b}}.
\newblock \href {https://aclanthology.org/2021.eacl-demos.22} {Massive choice,
  ample tasks ({M}a{C}h{A}mp): A toolkit for multi-task learning in {NLP}}.
\newblock In \emph{Proceedings of the 16th Conference of the European Chapter
  of the Association for Computational Linguistics: System Demonstrations},
  pages 176--197, Online. Association for Computational Linguistics.

\bibitem[{van~der Goot and van Noord(2017)}]{van2017monoise}
Rob van~der Goot and Gertjan van Noord. 2017.
\newblock Monoise: Modeling noise using a modular normalization system.
\newblock \emph{Computational Linguistics in the Netherlands Journal},
  7:129--144.

\bibitem[{Veliz et~al.(2019)Veliz, De~Clercq, and Hoste}]{veliz2019comparing}
Claudia~Matos Veliz, Orph{\'e}e De~Clercq, and V{\'e}ronique Hoste. 2019.
\newblock Comparing mt approaches for text normalization.
\newblock In \emph{Proceedings of the International Conference on Recent
  Advances in Natural Language Processing (RANLP 2019)}, pages 740--749.

\bibitem[{Wagner and Foster(2015)}]{wagner2015dcu}
Joachim Wagner and Jennifer Foster. 2015.
\newblock Dcu-adapt: learning edit operations for microblog normalisation with
  the generalised perceptron.
\newblock In \emph{Proceedings of the Workshop on Noisy User-generated Text},
  pages 93--98.

\bibitem[{Wang et~al.(2016)Wang, Liakata, Zubiaga, Procter, and
  Jensen}]{wang2016smiles}
Bo~Wang, Maria Liakata, Arkaitz Zubiaga, Rob Procter, and Eric Jensen. 2016.
\newblock Smile: Twitter emotion classification using domain.
\newblock In \emph{Proceedings of the 4th Workshop on Sentiment Analysis where
  AI meets Psychology co-located with 25th International Joint Conference on
  Artificial Intelligence}.

\bibitem[{Winata et~al.(2018)Winata, Kampman, and Fung}]{winata2018attention}
Genta~Indra Winata, Onno~Pepijn Kampman, and Pascale Fung. 2018.
\newblock Attention-based lstm for psychological stress detection from spoken
  language using distant supervision.
\newblock In \emph{2018 IEEE International Conference on Acoustics, Speech and
  Signal Processing (ICASSP)}, pages 6204--6208. IEEE.

\bibitem[{Wolf et~al.(2020)Wolf, Debut, Sanh, Chaumond, Delangue, Moi, Cistac,
  Rault, Louf, Funtowicz, Davison, Shleifer, von Platen, Ma, Jernite, Plu, Xu,
  Scao, Gugger, Drame, Lhoest, and Rush}]{wolf-etal-2020-transformers}
Thomas Wolf, Lysandre Debut, Victor Sanh, Julien Chaumond, Clement Delangue,
  Anthony Moi, Pierric Cistac, Tim Rault, Rémi Louf, Morgan Funtowicz, Joe
  Davison, Sam Shleifer, Patrick von Platen, Clara Ma, Yacine Jernite, Julien
  Plu, Canwen Xu, Teven~Le Scao, Sylvain Gugger, Mariama Drame, Quentin Lhoest,
  and Alexander~M. Rush. 2020.
\newblock \href {https://www.aclweb.org/anthology/2020.emnlp-demos.6}
  {Transformers: State-of-the-art natural language processing}.
\newblock In \emph{Proceedings of the 2020 Conference on Empirical Methods in
  Natural Language Processing: System Demonstrations}, pages 38--45, Online.
  Association for Computational Linguistics.

\bibitem[{Xu et~al.(2015)Xu, Xia, and Lee}]{xu2015tweet}
Ke~Xu, Yunqing Xia, and Chin-Hui Lee. 2015.
\newblock Tweet normalization with syllables.
\newblock In \emph{Proceedings of the 53rd Annual Meeting of the Association
  for Computational Linguistics and the 7th International Joint Conference on
  Natural Language Processing (Volume 1: Long Papers)}, pages 920--928.

\bibitem[{Yang and Srinivasan(2016)}]{yang2016life}
Chao Yang and Padmini Srinivasan. 2016.
\newblock Life satisfaction and the pursuit of happiness on twitter.
\newblock \emph{PloS one}, 11(3):e0150881.

\bibitem[{Yates et~al.(2017)Yates, Cohan, and Goharian}]{yates2017depression}
Andrew Yates, Arman Cohan, and Nazli Goharian. 2017.
\newblock Depression and self-harm risk assessment in online forums.
\newblock In \emph{Proceedings of the 2017 Conference on Empirical Methods in
  Natural Language Processing}, pages 2968--2978.

\bibitem[{Zampieri et~al.(2019{\natexlab{a}})Zampieri, Malmasi, Nakov,
  Rosenthal, Farra, and Kumar}]{zampieri-etal-2019-predicting}
Marcos Zampieri, Shervin Malmasi, Preslav Nakov, Sara Rosenthal, Noura Farra,
  and Ritesh Kumar. 2019{\natexlab{a}}.
\newblock \href {https://doi.org/10.18653/v1/N19-1144} {Predicting the type and
  target of offensive posts in social media}.
\newblock In \emph{Proceedings of the 2019 Conference of the North {A}merican
  Chapter of the Association for Computational Linguistics: Human Language
  Technologies, Volume 1 (Long and Short Papers)}, pages 1415--1420,
  Minneapolis, Minnesota. Association for Computational Linguistics.

\bibitem[{Zampieri et~al.(2019{\natexlab{b}})Zampieri, Malmasi, Nakov,
  Rosenthal, Farra, and Kumar}]{offenseval}
Marcos Zampieri, Shervin Malmasi, Preslav Nakov, Sara Rosenthal, Noura Farra,
  and Ritesh Kumar. 2019{\natexlab{b}}.
\newblock {SemEval-2019 Task 6: Identifying and Categorizing Offensive Language
  in Social Media (OffensEval)}.
\newblock In \emph{Proceedings of The 13th International Workshop on Semantic
  Evaluation (SemEval)}.

\bibitem[{Zampieri et~al.(2020)Zampieri, Nakov, Rosenthal, Atanasova,
  Karadzhov, Mubarak, Derczynski, Pitenis, and
  {\c{C}}{\"o}ltekin}]{zampieri-etal-2020-semeval}
Marcos Zampieri, Preslav Nakov, Sara Rosenthal, Pepa Atanasova, Georgi
  Karadzhov, Hamdy Mubarak, Leon Derczynski, Zeses Pitenis, and
  {\c{C}}a{\u{g}}r{\i} {\c{C}}{\"o}ltekin. 2020.
\newblock \href {https://aclanthology.org/2020.semeval-1.188} {{S}em{E}val-2020
  task 12: Multilingual offensive language identification in social media
  ({O}ffens{E}val 2020)}.
\newblock In \emph{Proceedings of the Fourteenth Workshop on Semantic
  Evaluation}, pages 1425--1447, Barcelona (online). International Committee
  for Computational Linguistics.

\end{thebibliography}
